\documentclass[runningheads, envcountsame, a4paper]{llncs}
\usepackage{booktabs}
\usepackage{color,soul}
\usepackage{xcolor,colortbl}
\usepackage{bm}
\usepackage{amsmath}
\usepackage{amssymb}
\usepackage{dsfont}
\definecolor{Gray}{gray}{0.9}
\usepackage{subcaption}
\usepackage{url}
\usepackage{graphicx}

\begin{document}
\title{Time series forecasting with Gaussian Processes needs priors}
\author{Giorgio Corani\inst{1} \and Alessio Benavoli\inst{2} \and Marco Zaffalon\inst{1}}
\authorrunning{Corani et al.}
\institute{Istituto Dalle Molle di Studi sull'Intelligenza Artificiale (IDSIA)\\
	USI - SUPSI\\
	Lugano, Switzerland\\
	\email{giorgio.corani\{marco.zaffalon\}@idsia.ch}
	\and
	School of Computer Science and Statistics\\ Trinity College Dublin\\ Ireland.
	\email{alessio.benavoli@tcd.ie}
}
\maketitle

\begin{abstract}
Automatic forecasting is the task of  receiving a time series and returning a forecast for the next time steps without any human intervention.
Gaussian Processes (GPs) are a powerful tool for
modeling time series, but so far there are no competitive approaches for automatic forecasting based on GPs.
We propose practical solutions to two problems: automatic selection of the optimal kernel and reliable estimation of the  hyperparameters.
We propose a fixed composition of kernels, which contains the components
needed to model most time series: linear trend, periodic patterns, and other flexible kernel for modeling the non-linear trend.
Not all components are necessary to model each time series; during training  the unnecessary components are automatically made irrelevant  via  automatic relevance determination (ARD).
We moreover assign priors to the hyperparameters, in order to keep the inference within a plausible range; we design such priors through an empirical Bayes approach.
We present results on many time series of different types; our GP model is more accurate than state-of-the-art time series models.
Thanks to the priors, a single restart is enough the estimate the hyperparameters; hence the model is also fast to train.
\end{abstract}

\section{Introduction}
Automatic forecasting \cite{pkg:forecast} is the task of
receiving a time series and returning a probabilistic forecast for the next time steps without any human intervention.
The algorithm should be both accurate and fast, in order to scale on a large number of time series,

Time series models such as exponential smoothing
(\textit{ets},  \cite{hyndman2018forecasting}) and automated arima procedures
(\textit{auto.arima} \cite{pkg:forecast}) are strong baselines
on monthly and quarterly time series, which contain limited number of samples.
In these cases they generally outperform recurrent neural networks  \cite{HEWAMALAGE2020},
which are also much more time-consuming to train.

Time series which are sampled at higher frequency
generally contain multiple seasonal patterns. For instance, a time series of hourly data
typically contains a  daily and a weekly seasonal pattern.
This type of time series can be forecasted with models
such as \textit{tbats} \cite{de2011forecasting} and \textit{Prophet} \cite{taylor2018forecasting}.

Gaussian Processes (GPs) \cite{rasmussen2006gaussian}
are a powerful tool for modeling correlated observations, including time series.
The GP provides a prior over functions, which  captures prior beliefs about the function behavior, such as smoothness or periodicity. Given some observations, the prior is updated to form the posterior distribution over the  functions. Dealing with Gaussian noise, this posterior distribution is again a GP. The posterior GP is used to predict the value of the function in points which have yet to be sampled; this  prediction is accompanied by a principled quantification of the uncertainty.
GPs have been used for the
analysis of astronomical time series (see \cite{foreman2017fast} and the references therein),
forecasting  of electric load \cite{LLOYD2014369} and
analysis of correlated and irregularly-sampled  time series \cite{roberts2013gaussian}.

Within a GP model,
the kernel determines which functions are  used for curve fitting.
Complex functions can be obtained by summing or multiplying
basic kernels; this is called
\textit{kernel composition}.
In some cases the composition can be based on physical considerations
\cite{foreman2017fast} or personal expertise
\cite{LLOYD2014369}. However
algorithms which
automatically optimize the kernel composition
\cite{duvenaud2013structure,pmlr-v64-malkomes_bayesian_2016,kim2018scaling} do not scale,
given the need for training a large number of competing GP models, each with cubic complexity.
Moreover, there is no result showing that this strategy can forecast as accurately as
the best time series models.

Summing up, there are currently  no competitive approaches for automatic forecasting based on Gaussian Processes. In this paper we fill this gap, proposing a GP model which is accurate, fast to train and suitable for different types of time series.

We propose a kernel composition  which contains useful components for modeling time series:
linear trend, periodic patterns, and other flexible kernel for modeling the non-linear trend.
We keep this composition fixed, thus avoiding kernel search.
When dealing with a specific time series, some components might be unnecessary;
during training
they are made automatically  irrelevant by
\textit{automatic relevance determination} (ARD) \cite{mackay1998introduction}.
Indeed, ARD yields automatic feature selection for GPs.

We then consider how to reliably estimate the hyperparameters even on short time series.
We keep their inference
within a reasonable range by assigning priors to them.
We  define the parameters of such priors  by means of a Bayesian hierarchical model trained on a separate subset of time series.

Extensive results show that
our model is very accurate and versatile. It generally outperforms the state-of-the-art competitors on
monthly and quarterly time series; moreover, it can be easily extended to model
time series with double seasonality. Also in this case, it
compares favorably to specialized time series models.
A single restart is enough to sensibly estimate the hyperparameters; hence the model is also fast to train.

The paper is organized as follows: in Sec.\ref{sec:GP} we introduce GPs;
in Sec.\ref{sec:composition} we present our kernel composition and the definition of the  priors;
in Sec. \ref{sec:expe} we present the experiments.

\section{Gaussian processes}\label{sec:GP}
We cast time-series modelling as a regression problem:
\begin{equation}
\label{eq:model}
y=f(\mathbf{x})+v,
\end{equation}
where $\mathbf{x} \in  \mathbb{R}^p$,  $f : \mathbb{R}^p \rightarrow \mathbb{R}$
and $v\sim N(0,s_v^2)$
is the noise.
We assume
a Gaussian Process (GP) as a prior distribution about function $f$:
\begin{equation*}
	f \sim GP(0,k_{\bm{\theta}}),
\end{equation*}
where
$k_{\bm{\theta}}$ denotes the kernel  with hyperparameters $\bm{\theta}$.
It is common to adopt
the
zero function as a mean function, since a priori we do not know whether at any point the trend will be below or above the average \cite{roberts2013gaussian}.

The kernel defines the covariance between the value of the function in different locations:
 $Cov(f(\mathbf{x}),  f(\mathbf{x}^*) ) = k_{\bm{\theta}}(\mathbf{x}, \mathbf{x}^* )$, $k_{\bm{\theta}}:\mathbb{R}^p\times \mathbb{R}^p  \rightarrow \mathbb{R}^+$
and thus it determines which functions are likely under the GP prior.

The most common kernel is the \textit{squared exponential}, also referred to as radial basis funcion (RBF):
\begin{equation*}
	\text{RBF}:~ \displaystyle k_{\bm{\theta}}(x_1, x_2) = s_r^2 \exp \left(-\frac{(x_1-x_2)^2}{2 \ell_r^2}\right),\\
\end{equation*}
whose hyperparameters are the variance $s_r^2$ and the lengthscale
$\ell_r$.
Longer lengthscales yields smoother functions and
shorter lengthscales yields wigglier functions.
A limit of the  RBF kernel is that, once conditioned on the training data, it does not extrapolate more than $\ell$ units away from the observations.

The periodic (PER) kernels yields
periodic functions which repeat themselves exactly.
Such function correspond to the sum of infinite  Fourier terms \cite{solin2014explicit,benavoli2016state} and hence the PER kernel
can represent any periodic function.
It is defined as:
\begin{equation*}
	\text{PER:}~ \displaystyle  k_{\bm{\theta}}(x_1, x_2) = s_p^2 \exp
	\left(-\frac{(2 \sin^2(\pi|x_1-x_2|/p_e)}{\ell_p^2}\right),\\
\end{equation*}
where $\ell_p^2$ controls the wiggliness of the functions, $p_e$ denotes the period and
$s_p^2$ the variance.

Notice that in general, when the lengthscale  of a kernel
tends to infinity,  or its variance tends to zero, the
kernel yields functions that vary less and less as a function of $x$.

The linear kernel, which yields linear functions, is:
\begin{equation*}
	\text{LIN}:    k_{\bm{\theta}}(x_1, x_2) = s_b^2 + s_l^2 x_1x_2,\\
\end{equation*}
A GP with LIN kernel is equivalent \cite{rasmussen2006gaussian} to a Bayesian linear regression.

The white noise (WN) kernel, which is used to represent the noise of the regression, is:
\begin{equation*}
	\text{WN}: k_{\bm{\theta}}(x_1, x_2) = s_v^2 \delta_{x_1,x_2}.
\end{equation*}

The above expressions are valid for $p=1$, which is the case of a univariate time series;  see \cite{rasmussen2006gaussian} for  the case $p$\textgreater1  and further kernels.

\subsection{Kernel compositions}
Positive definite kernels (i.e., those which define valid covariance functions) are closed under addition and multiplication \cite{rasmussen2006gaussian}.
Hence, complex functions can be modeled by adding or multiplying simpler kernels; this is called composition.

There are algorithms which iteratively train and compare GPs equipped with different
kernel compositions
\cite{duvenaud2013structure,pmlr-v64-malkomes_bayesian_2016},
but they are characterized by large computational complexity.
Even if recent works have made the procedures more scalable
\cite{kim2018scaling,teng2020scalable},
they are still not comparable to lighting-fast time series model.

The \textit{spectral mixture} kernel \cite{wilson2013gaussian}
allows the GP to fit
complex functions without
kernel search.
It is defined as the sum of Q components, where the $i$-th component
is:
\begin{equation*}
	\text{SM}_i: k_{\bm{\theta}}(x_1, x_2) = s_{m_i}^2 \exp \left(-\frac{(x_1-x_2)^2}{2 \ell_{m_i}^2}\right)
	\cos \left(   \frac{x_1-x_2}{\tau_{m_i}}  \right), \label{eq:sm}\\
\end{equation*}
with hyperparameters are $s_{m_i}$,  $\ell_{m_i}$ and $\tau_{m_i}$.
It also corresponds to the product of a RBF kernel and another kernel called cosine kernel.
Estimating the hyper-parameters  of the SM kernel is however challenging:
the marginal likelihood is highly multimodal and it is unclear how
to initialize the optimization.
In \cite{wu2017bayesian} Bayesian optimization is used for deciding the
initialization at each restart. This is effective but requires quite a few restarts.

\subsection{The  composition}\label{sec:composition}
We propose the following kernel composition:
\begin{equation}
	\text{K = PER + LIN + RBF} + \text{SM}_1 + \text{SM}_2,
	\label{eq:kern-comp}
\end{equation}
which arguably contains the most important components for forecasting.

The  periodic kernel (PER) models the seasonal pattern; for monthly and quarterly time series,
we assume a  period of one year and we set $p_e$=1.
Time series with a double
seasonality can be modeled by adding a second
periodic kernel,
as we do in Sec. \ref{sec:energy}.

The LIN kernel provides the linear trend.  This is an  important component: for instance, auto.arima \cite{pkg:forecast} adds a linear trend
(by applying first differences) to about 40\% of the monthly time series of the M3 competition.
The RBF and the two SM kernels are intended to model non-linear trends which might characterize the time series.

\subsubsection{Automatic Relevance Determination}
Some components of the composition might be unnecessary when fitting
a certain time series: for instance,
a time series might show no seasonal pattern or no linear trend.
This is automatically managed via
\textit{automatic relevance determination} (ARD) \cite{mackay1998introduction}.
When fitting the hyperparameters, the unnecessary components are given long lengthscale and/or small variance; in this way they are made irrelevant within the curve being fitted.

\subsection{Training strategy}
Reliably estimating the hyperparameters of the GP can be challenging
(see e.g. \cite{wu2017bayesian}), especially when dealing with small data sets such as
monthly and quarterly time.

We keep the inference of the hyperparameters within a plausible range by assigning priors to them.
Variances and lengthscales are non-negative parameters, to which we assign
log-normal priors:
\begin{align}                                                                               	& s^{2}_l,s^{2}_r,s^{2}_p,s^{2}_{m_1},s^{2}_{m_2}, s^{2}_v\sim \text{LogN}(\nu_s,\lambda_s)   \label{eq:logn-var}\\
	& \ell_{r} \sim \text{LogN}(\nu_r,\lambda_{\ell})\\
	& \ell_{p} \sim \text{LogN}(\nu_p,\lambda_{\ell})\\
	& \ell_{m_1} \sim \text{LogN}(\nu_{m_1}, \lambda_{\ell})\\
	& \ell_{m_2} \sim \text{LogN}(\nu_{m_2},\lambda_{\ell})\\
	& \tau_{m_1} \sim \text{LogN}(\nu_{t_1}, \lambda_{\ell})\\
	& \tau_{m_2} \sim \text{LogN}(\nu_{t_2},\lambda_{\ell}),
	\label{eq:logn-sm2}
\end{align}
where $\text{LogN}(\nu,\lambda)$ denotes the distribution with mean $\nu$ and variance $\lambda$.

According to Eq.\eqref{eq:logn-var},
all components share the same prior on the variance.
This assign to every component the same prior probability of being irrelevant,
as a component can be made irrelevant by pushing
its variance to zero.
We assign moreover a shared variance $\lambda_{\ell}$ to all lengthscales, in order to
simplify the  numerical fitting of the hierarchical model described in the next section.

We manage time such that time increases of one unit when one year has passed.
The lengthscales can be readily interpreted; for instance an RBF kernel with lengthscale
of 1.5 years is able to forecast about 1.5 years in the future before reverting to the prior mean.

\subsubsection{Hierarchical GP model}
\label{sec:hierGP}
To numerically define the priors \eqref{eq:logn-var}--\eqref{eq:logn-sm2}, we
adopt an empirical Bayes approach.
We select a set of $B$ time series and we fit a \textit{hierarchical} GP model to extract distributional information about the hyperparameters.
The
hierarchical Bayes model allows learning different models from
different related data sets \cite[Chap. 5]{bda2013}.
Example of hierarchical GP models,  not  related to time series,  are given in
\cite{lawrence2004learning} and
\cite{schwaighofer2005learning}.

We assume the
hyperparameters of the different time series
to be drawn from  higher-level priors (\textit{hyperprior}).
For instance the
lengthscales of the RBF kernel ($\ell^{(1)}_{r},\ell^{(2)}_{r},..., \ell^{(B)}_{r} $)
are all drawn from the same hyperprior.

The generative model for
the j-th time series is hence:
$$
\begin{aligned}
	& s^{2(j)}_l,s^{2(j)}_r,s^{2(j)}_p,s^{2(j)}_{m_1},s^{2(j)}_{m_2}, s^{2(j)}_v\sim \text{LogN}(\nu_s,\lambda_s)\\
	&\ell^{(j)}_{r} \sim \text{LogN}(\nu_r,\lambda_{\ell})\\
	& \ell^{(j)}_{p} \sim \text{LogN}(\nu_p,\lambda_l),\\
	&\ell^{(j)}_{m_1} \sim \text{LogN}(\nu_{m_1}, \lambda_{\ell}) \\
	& \ell^{(j)}_{m_2} \sim \text{LogN}(\nu_{m_2},\lambda_{\ell}),\\
	&\tau^{(j)}_{m_1} \sim \text{LogN}(\nu_{\tau_1}, \lambda_{\ell}) \\
	&\tau^{(j)}_{m_2} \sim \text{LogN}(\nu_{\tau_2},\lambda_{\ell}),\\
	&\mathbf{y}^{(j)}\sim N(0, K_{\bm{\theta}}^{(j)}(X^{(j)},X^{(j)})),
\end{aligned}
$$
where  $K$  denotes our kernel composition, instantiated with the hyper-parameters of the j-th time series;
$\bm{\theta}^{(j)}$ denotes
the  hyper-parameters of the j-th time series.

We assign weakly-informative priors to the $\nu,\lambda$  parameters:
\begin{align}
	& \nu_s,\nu_p,\nu_r,\nu_{m_2} \sim N(0,5) \label{eq:prior-hier}\\
	& \nu_{m_1} \sim N(-1.5,5)\\
	& \lambda_s, \lambda_l \sim \text{Gamma}(1,1).\label{eq:prior-hier2}
\end{align}
The lower prior mean for $\nu_{m_1}$ is helpful for differentiating the estimation of
SM$_1$ and SM$_2$ towards shorter-term and longer-term trends respectively.

\begin{figure}[!htp]
	\centering
	\includegraphics[width=0.49\linewidth]{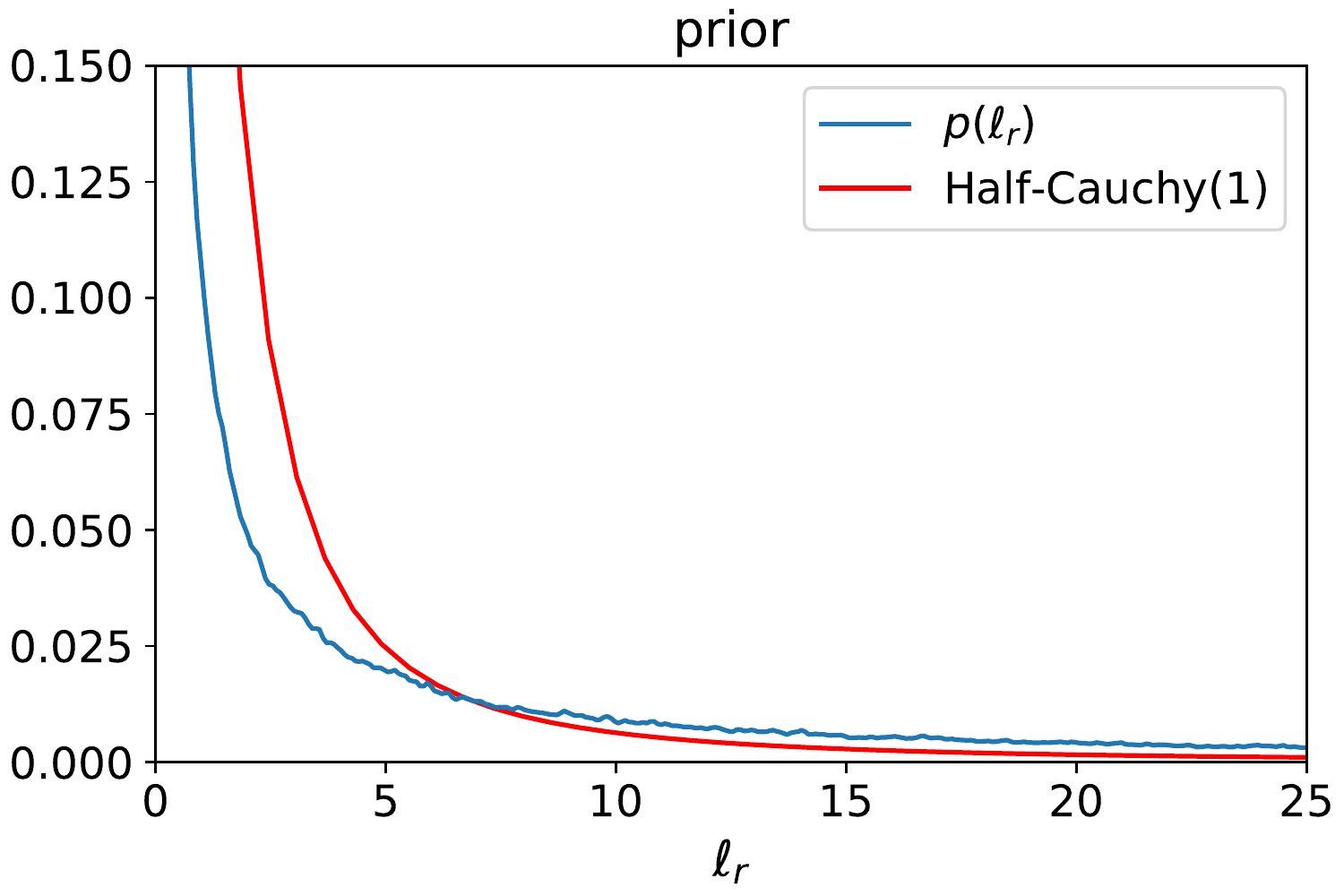}
	\includegraphics[width=0.49\linewidth]{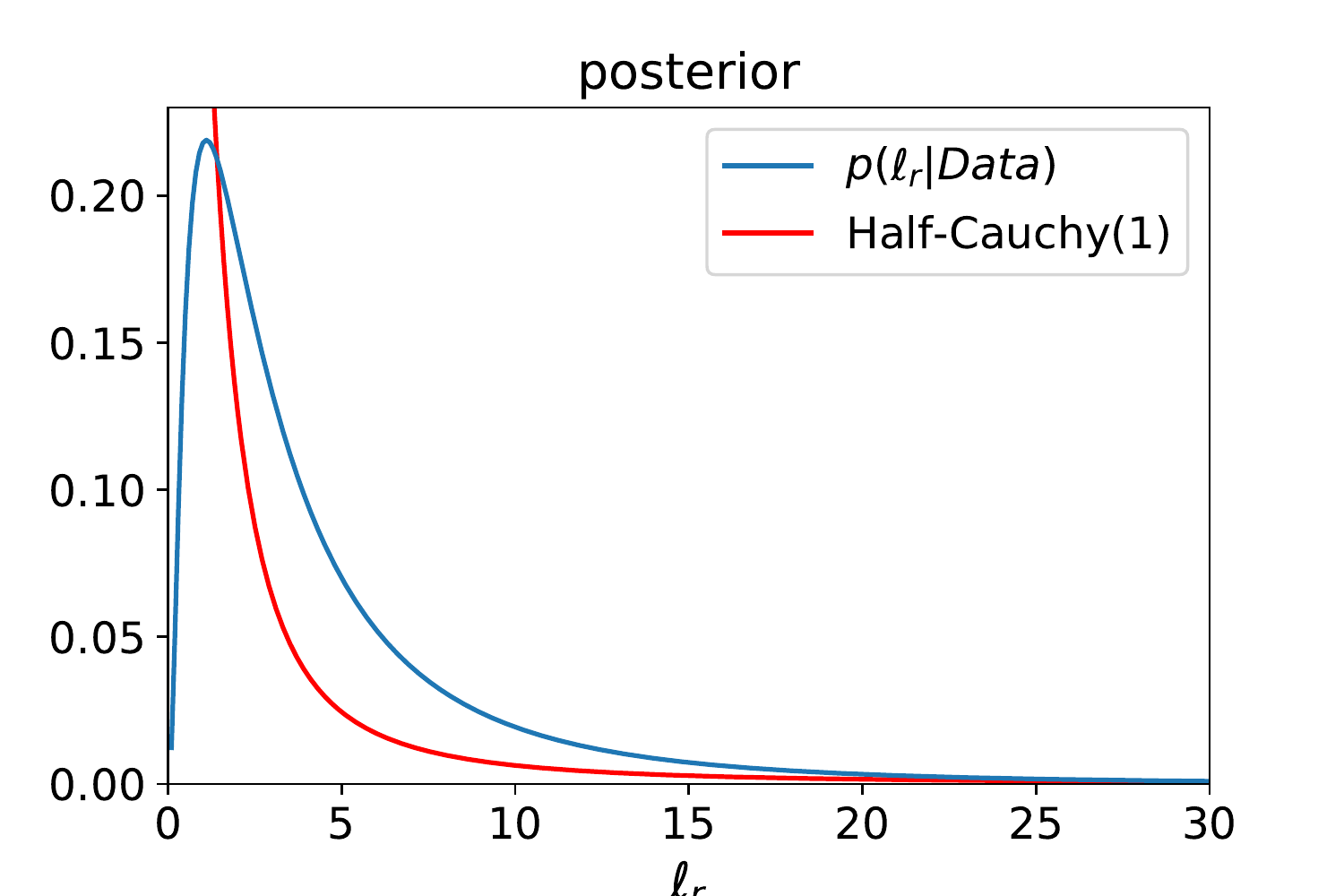}
	\caption{Left: prior on $\ell_r$ induced by the hierarchical model. Right:
		posterior on $\ell_r$ estimated by the hierarchical model using 350 time-series. The Half-Cauchy distribution (with scale=1) is shown for comparison.
		In this paper we represent
		time such that,
		when one year has passed,
		$x$ increases of one unit.
	}
	\label{fig:hyper}
\end{figure}

We implemented the model in PyMC3 \cite{salvatier2016probabilistic}.
We use automatic differentiation variational inference to approximate the posterior distribution of the $\nu$'s and $\lambda$'s.
We fit the hierarchical model on 350 monthly time series from the M3 competition.
Before fitting the hierarchical model, we standardize each time series to have mean 0 and variance 1. Moreover, we manage time such that time increases of one unit when one year has passed.

The priors induced by the hierarchical model have fat tails.
Consider for instance
the prior induced on $\ell_r$, which according to Eq.\eqref{eq:prior-hier}
-- \eqref{eq:prior-hier2} is:
$p(\ell_r)=\iint \text{LogN}(\ell_r;\nu_r,\lambda_{\ell})N(\nu_r;0,5)\text{Gamma}(\lambda_l;1,1) d\nu_r d\lambda_l$.
It is shown in the left plot of
Fig.\ref{fig:hyper},
and its tails are actually fatter than those
of the Half-Cauchy distribution.

Figure \ref{fig:hyper}(right) shows instead the distribution on $\ell_r$ obtained
using the  posterior means of $\nu_r$ and $\lambda_l$, estimated by the hierarchical model.  This yields a distribution on $\ell_r$ which we use
as prior  when fitting the GP.
This prior has fat tails too, see the comparison with the half-Cauchy;  nevertheless, it does inform the optimizer about the order of magnitude of $\ell_r$.
The median and the 95-th percentile of the prior of each hyperparameter are given in Tab.\ref{tab:priors}.

\begin{table}[!htp]
	\setlength{\tabcolsep}{5pt}
	\centering
	\begin{tabular}{@{}lrrlrr}
		\toprule
		parameter                      & median & 95th & $\,\,\,\,\,$ & $\,\,\,\,\nu\,\,\,\,$ & $\,\,\,\,\lambda\,\,\,\,$ \\ \midrule
		\rowcolor{Gray}
		variance     &    0.2 &  1.2 & & -1.5                  & 1.0                       \\
		std\_periodic                  &    1.2 &  6.3 & & 0.2                   & 1.0                       \\
		\rowcolor{Gray}
		rbf          &    3.0 & 15.4 & & 1.1                   & 1.0                       \\
		SM$_1$ (rbf)                   &    0.5 &  2.5 & &    -0.7                   & 1.0                           \\
		\rowcolor{Gray}
		SM$_1$ (cos) &    1.7 &  8.6 & &                 0.5      & 1.0                           \\
		SM$_2$ (rbf)                   &    3.0 & 15.4 & &             1.1          & 1.0                           \\
		\rowcolor{Gray}
		SM$_2$ (cos) &    5.0 & 25.8 & &                 1.6    & 1.0                            \\ \bottomrule
		                               &        &      & &                       &
	\end{tabular}
	\caption{Quantiles on the hyperparameters implied by the lognormal priors and parameters ($\nu, \lambda$) of the lognormal priors. By design, the $\lambda$s are equal for all lengthscales.}
	\label{tab:priors}
\end{table}

The prior on the variance is coherent with the fact that we work with standardized time series, whose variance is one.

The priors over the lengthscales also yield plausible ranges, every component having a median lengthscale
comprised  between 0.5 and 3 years, with long tails arriving up to 25 years.

All the experiments of this paper are thus computed using the priors of Tab.\ref{tab:priors}.
To remove any danger of overfitting, we remove the 350 time series used to fit the hierarchical model from our experiments.

\paragraph{Further considerations}
%

In the jargon of time series, models which are fitted to a set of time series
are referred to as \textit{global} models, see for instance \cite{MONTEROMANSO202086,salinas2020deepar}. The hierarchical model is a global model,
as it jointly analyzes different time series.
Global models can be more accurate than univariate models, if the time series are characterized by some common patterns. Yet, they are also more complicated to fit.
In this paper we do not consider global models.
We use the hierarchical model only for defining the priors on
the hyperparameters of the GP.

\subsection{MAP estimation}
We estimate the
hyperparameters by
computing the maximum a-posteriori (MAP) estimate  of  $\bm{\theta}$,
thus  approximating the marginal of $\mathbf{f}^*$ with \eqref{eq:post}.
We thus
maximize  w.r.t.\  $\bm{\theta}$  the joint marginal probability of $\mathbf{y},\bm{\theta}$, which is the product of the prior $p(\bm{\theta})$ and
the marginal likelihood  \cite[Ch.2]{rasmussen2006gaussian}:
\begin{equation}
	\label{eq:mllike}
	\begin{array}{ll}
		p(\mathbf{y}|X,\bm{\theta})=& N(\mathbf{y};0, K_{\bm{\theta}}(X,X)).\end{array}
\end{equation}
Using a single restart, MAP estimation is generally accomplished in less than a second (on a standard computer) on monthly and quarterly time series, yielding thus quick training times.

\subsection{Forecasting}
Based on the training data $X^T=[\mathbf{x}_1,\dots,\mathbf{x}_n]$, $\mathbf{y}=[y_1,\dots,y_n]^T$ , and given $m$ test inputs $(X^*)^T=[\mathbf{x}_1^*,\dots,\mathbf{x}_m^*]$ , we wish to find the posterior
distribution of $\mathbf{f}^*= [f(\mathbf{x}^*_1),\dots,f(\mathbf{x}^*_m)]^T$.

\begin{figure}[!h]
	\centering
	\begin{subfigure}[b]{0.49\linewidth}
		\centering
		\includegraphics[width=\linewidth]{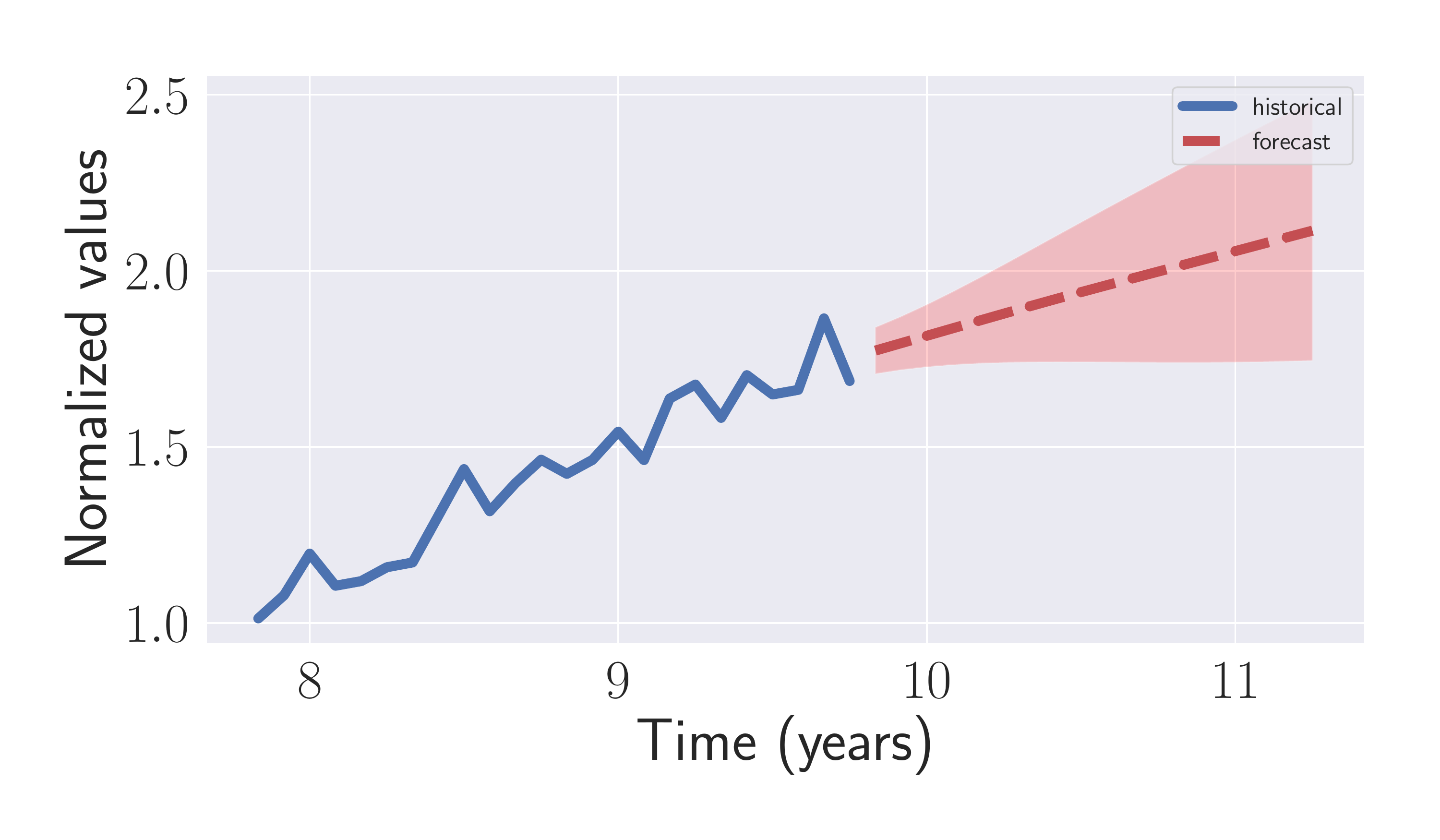}
	\end{subfigure}
	\begin{subfigure}[b]{0.49\linewidth}
		\centering
		\includegraphics[width=\linewidth]{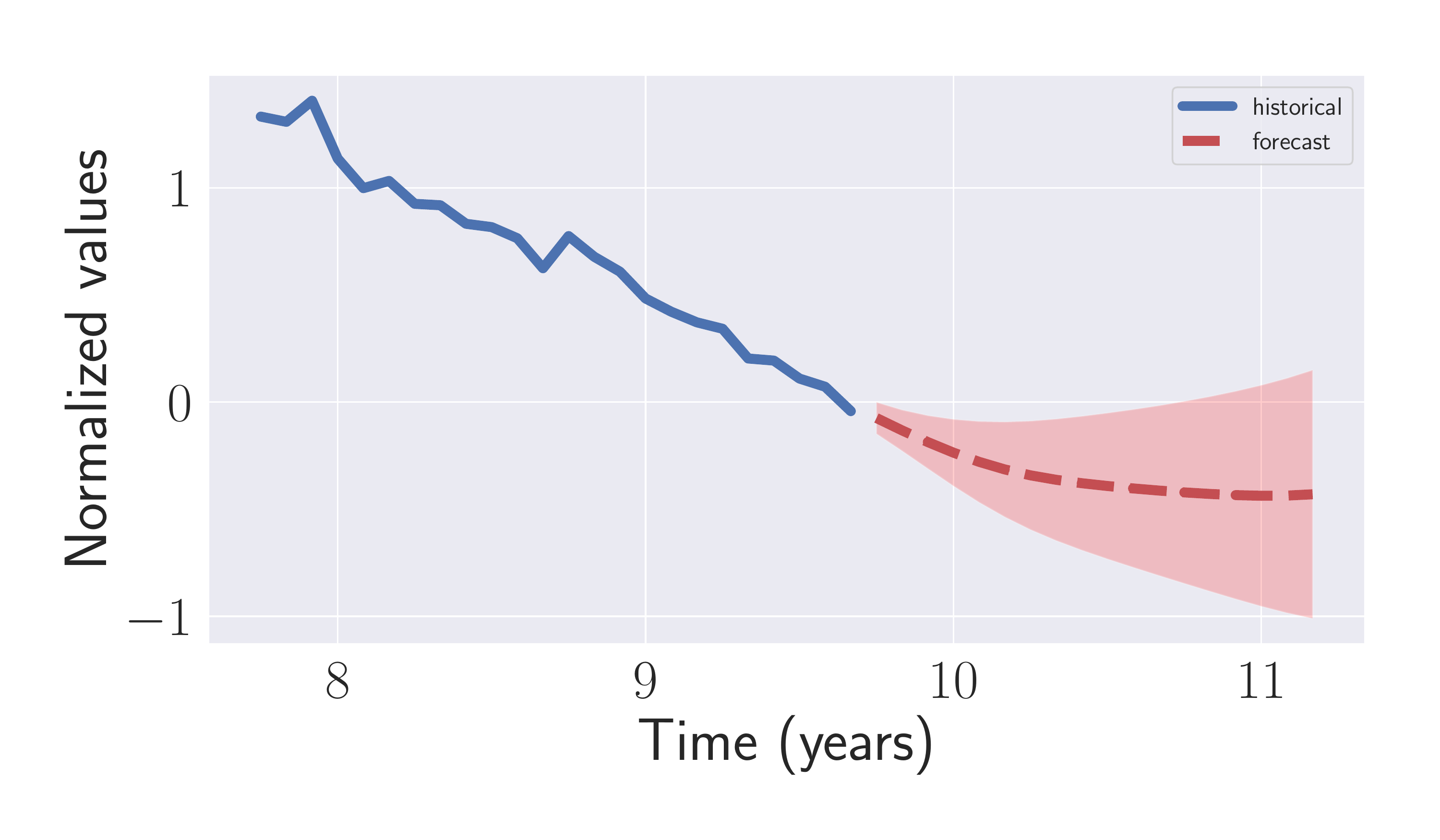}
	\end{subfigure}
	\begin{subfigure}[b]{0.49\linewidth}
		\centering
		\includegraphics[width=\linewidth]{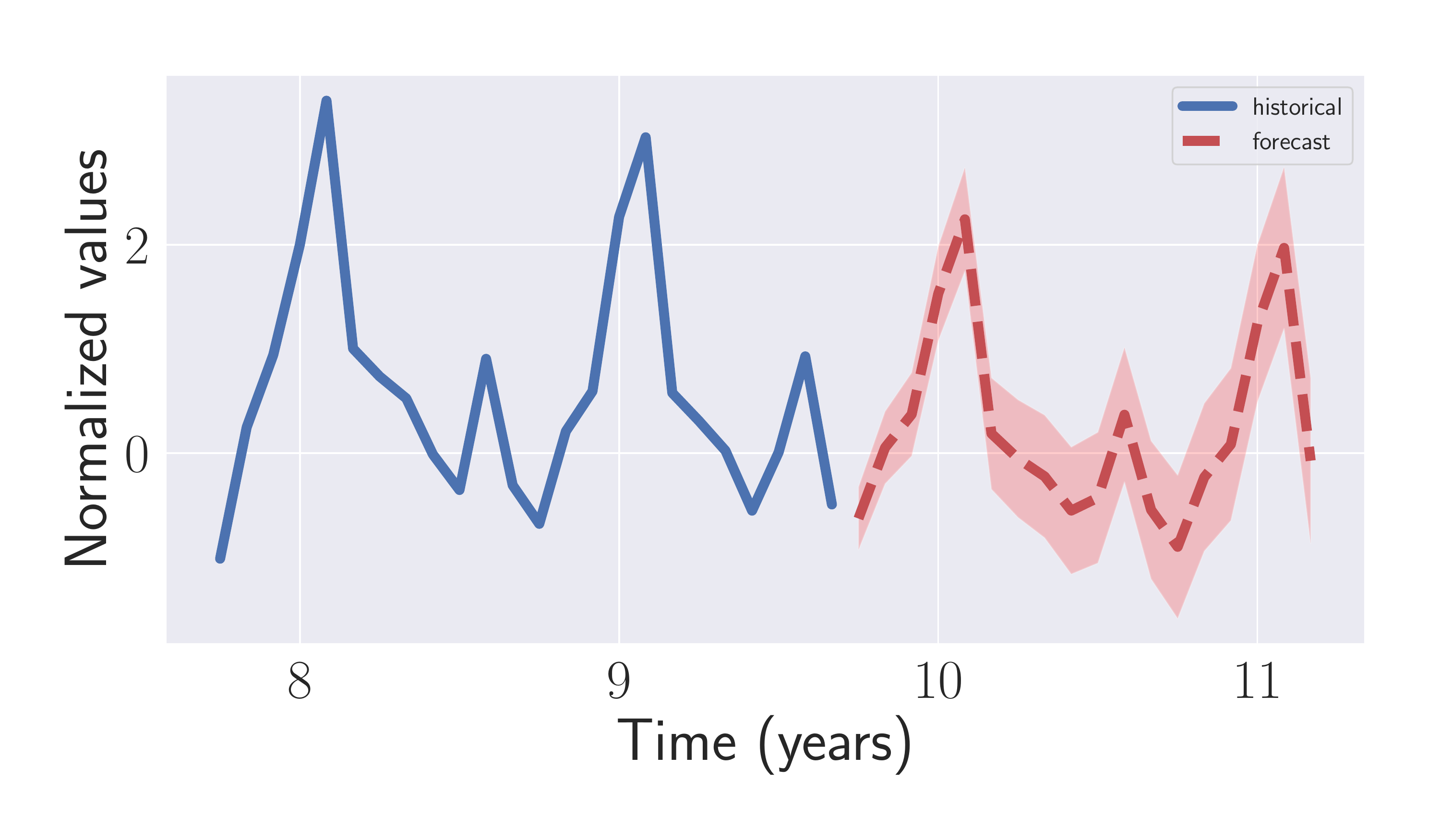}
	\end{subfigure}
	\begin{subfigure}[b]{0.49\linewidth}
		\centering
		\includegraphics[width=\linewidth]{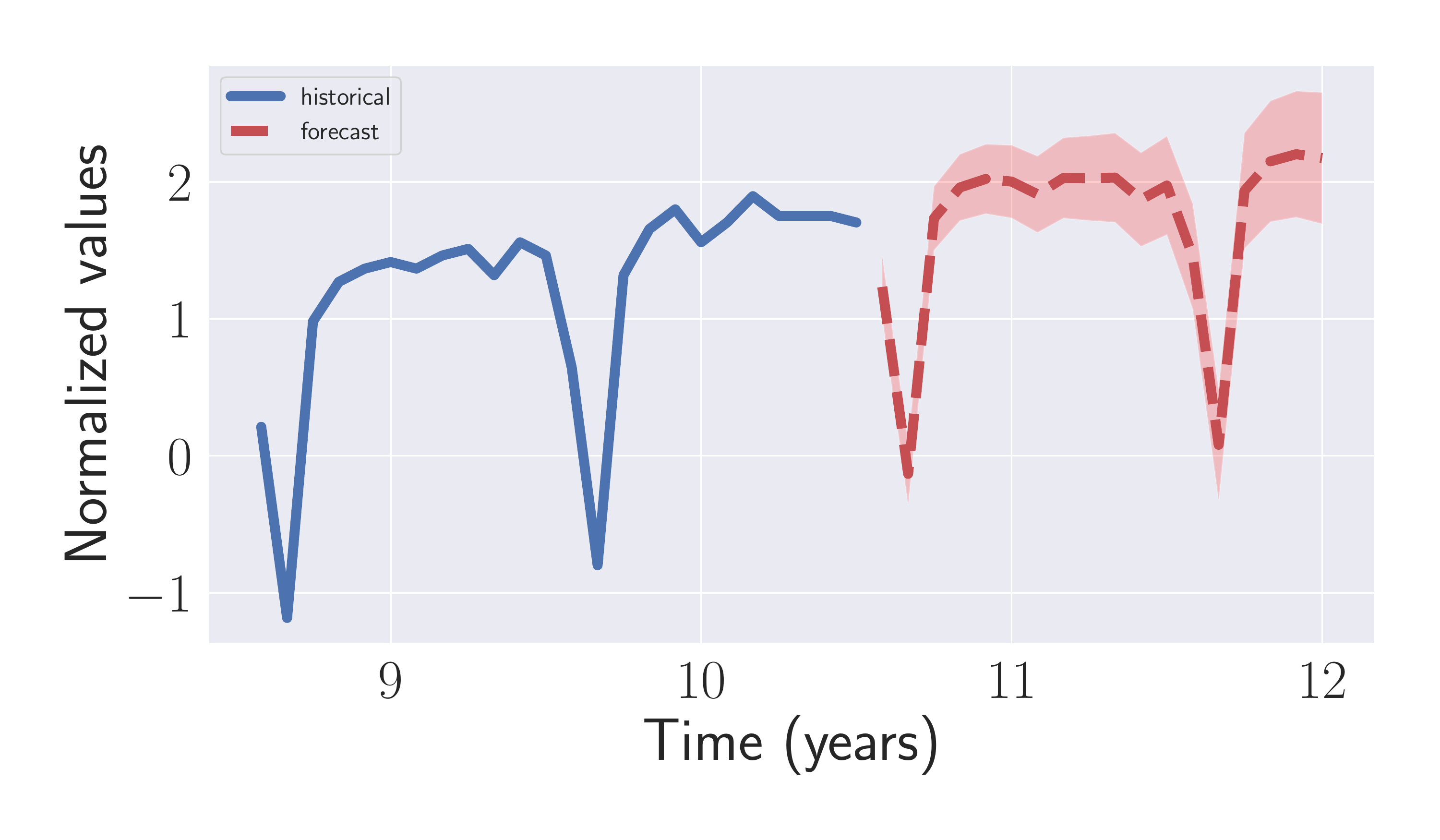}
	\end{subfigure}
	\caption{
		Examples of GP forecasts on monthly time series, computed up to 18 months ahead.}
	\label{fig:GPpatterns}
\end{figure}

From (\ref{eq:model}) and the properties of the Gaussian distribution, \footnote{In the paper, we  incorporate  the additive  noise $v$  into  the  kernel by adding a White noise kernel term.}
the posterior distribution of $\mathbf{f}^*$ is \cite[Sec. 2.2]{rasmussen2006gaussian}:
\begin{equation}
		\label{eq:post}
		p(\mathbf{f}^*|X^*,X,\mathbf{y},{\bm{\theta}}) = N(\mathbf{f}^*; \hat{\boldsymbol{\mu}}_{\bm{\theta}}(X^*|X,\mathbf{y}),\hat{K}_{\bm{\theta}}(X^*,X^*|X)),
\end{equation}
with mean and covariance given by:
\begin{align}
	\nonumber
	&\hat{\boldsymbol{\mu}}_{\bm{\theta}}(\mathbf{f}^*|X,\mathbf{y})=K_{\bm{\theta}}(X^*,X)(K_{\bm{\theta}}(X,X))^{-1}\mathbf{y},\\
	&\hat{K}_{\bm{\theta}}(X^*,X^*|X)=K_{\bm{\theta}}(X^*,X^*)\\
	\nonumber
	&-K_{\bm{\theta}}(X^*,X)(K_{\bm{\theta}}(X,X))^{-1}K_{\bm{\theta}}(X,X^*).
\end{align}

Our kernel composition, trained using the proposed priors,  yields sensible forecasts in very different contexts, as in Fig.\ref{fig:GPpatterns}.

\section{Experiments}\label{sec:expe}
We run experiments on the monthly and quarterly time
series of the
M1 and  M3 competitions, available from the package
Mcomp \cite{pkg:Mcomp} for R.
The original 1428 time series of the M3 competition drop to 1078 once we remove the 350 time series used to fit the hierarchical model.
Overall we consider about 959 quarterly time series (203 from M1, 756 from M3)
and 1695 monthly time series (617 from M1 and 1078 from M3).
The test set of monthly time series contains 18 months;
the test set of quarterly time series contains 8 quarters.
We standardize
each time series to have mean 0 and variance 1 on the training set.

\renewcommand\arraystretch{1.5}
\begin{table}[!htp]
	\setlength{\tabcolsep}{10pt}
	\centering
	\begin{tabular}{@{}crrrr}
		\toprule
		&                      \multicolumn{2}{c}{quarterly}                       &                       \multicolumn{2}{c}{monthly}                        \\
		& \multicolumn{1}{c}{M1} & \multicolumn{1}{c}{M3} & \multicolumn{1}{c}{M1} & \multicolumn{1}{c}{M3} \\ \midrule
		number of time series  &                    203    &                     756   &                     617   &       1078                 \\
		\rowcolor{Gray}
		median training length             &                  40      &         44               &               66         &                115        \\
		Test set length           &                      8 &                      8 &                     18 &                     18 \\ \bottomrule
	\end{tabular}
	\caption{Main characteristics of the M1 and M3 data sets.}
	\label{tab:M-competition}
\end{table}

We denote by GP our model trained \textit{with} priors and by GP$_0$
our model trained \textit{without} priors, i.e.,  by maximizing the marginal likelihood.
We use a single restart when training both GP and GP$_0$; on these time series, which
contain around 100 observations, the average training is generally less than one second on a standard laptop.

As competitors we consider \textit{auto.arima} and  \textit{ets}, both available from
the forecast package \cite{hyndman2018forecasting} for R.
We tried also Prophet \cite{taylor2018forecasting}, but its accuracy was not competitive. We thus dropped it;  we will consider it later in experiments with different types of time series.

\renewcommand\arraystretch{1.5}
\setlength{\tabcolsep}{7pt}
\begin{table}[!htp]
	\centering
	\begin{tabular}{@{}cccrrrr@{}}
		\toprule
		competition & freq      & score & GP             & \multicolumn{1}{c}{\textit{ets}}            & \textit{\small{arima}}          &
		\multicolumn{1}{c}{GP$_0$}   \\ \midrule
		\rowcolor{Gray}
		M1          & monthly   & MAE   & \textbf{0.58}  & 0.59$\,\,$           & 0.62$^*$           & 0.72$^*$  \\
		\rowcolor{Gray}
		M1          & monthly   & CRPS  & \textbf{0.41}  & 0.45$^*$           & 0.45$^*$           & 0.53$^*$  \\
		\rowcolor{Gray}
		M1          & monthly   & LL    & \textbf{-1.13} & -1.27$^*$          & -1.28$^*$          & -1.67$^*$ \\
		M1          & quarterly & MAE   & \textbf{0.57}  & 0.63$^*$           & 0.62$^*$          & 0.75$^*$  \\
		M1          & quarterly & CRPS  & \textbf{0.39}  & 0.47$^*$           & 0.44$^*$           & 0.59$^*$  \\
		M1          & quarterly & LL    & \textbf{-1.07} & -1.41$^*$          & -1.44$^*$         & -2.66$^*$ \\
		\rowcolor{Gray}
		M3          & monthly   & MAE   & \textbf{0.48}  & 0.51$^*$           & 0.51$^*$           & 0.59$^*$  \\
		\rowcolor{Gray}
		M3          & monthly   & CRPS  & \textbf{0.35}  & 0.38$^*$           & 0.37$^*$           & 0.42$^*$  \\
		\rowcolor{Gray}
		M3          & monthly   & LL    & \textbf{-1.01} & -1.05$^*$          & -1.06$^*$          & -1.23$^*$ \\
		M3          & quarterly & MAE   & 0.42$\,\,$           & \textbf{0.41}$\,\,$           & \textbf{0.41}$\,\,$  & 0.54$^*$  \\
		M3          & quarterly & CRPS  & \textbf{0.30}$\,\,$  & 0.31$\,\,$           & 0.31$\,\,$  & 0.40$^*$  \\
		M3          & quarterly & LL    & \textbf{-0.85} & -0.90$^*$          & -0.94$^*$          & -1.61$^*$ \\
\bottomrule
	\end{tabular}
	\caption{Median results on M1 and M3 time series.The best-performing model is boldfaced. 	Starred results correspond to the GP yielding a  significant improvement over the competitor (95\%,  Bayesian signed-rank test).}
	\label{tab:results-m}
\end{table}

\subsection*{Indicators}
Let us denote by $y_t$ and $\hat{y}_t$ the actual and the expected value
of the time series at time $t$; by $\sigma^2_t$ the variance of the forecast at time $t$;
by $\text{T}$ the length of the test set.
The mean absolute error (MAE)  on the test set is:
\begin{align*}
& \text{MAE}= \sum_{t=1}^{T} |y_t - \hat{y}_t|\\
\end{align*}

The continuous-ranked probability score (CRPS) \cite{gneiting2007strictly} is a proper scoring rule which generalizes MAE to probabilistic forecasts.
Let us denote by $F_t$ the cumulative predictive distribution  at time $t$ and
and by $z$ the variable over which we integrate.
The CRPS is:

\begin{align*}
	\text{CRPS}(F_t, y_t)=-\int_{-\infty}^{\infty}(F_t(z)- \mathds{1} \{z \geq y_t\})^{2} \mathrm{d}z.
	\label{eq:crps_integral}
\end{align*}

The log-likelihood of the test set (LL) is defined as:
\begin{align*}
& \text{LL}= \frac{1}{T}\left(
-\frac{\,1\,}{2} \sum_{t=1}^T \log(2\pi\sigma^2_t)
- \frac{1}{2\sigma^2_t} \sum_{t=1}^T (\, y_t - \hat{y}_t \,)^2 \right)
\end{align*}

MAE and CRPS are loss functions, hence the lower the better;
instead for LL, the higher the better.

In Tab.~\ref{tab:results-m} we report the median results for each indicator and each data set.
In each setting the GP yields the best median on almost all indicators.
However GP$_0$ is instead clearly outperformed by both ets and auto.arima.
Hence, our GP  model needs priors on the hyperparameters to produce
highly accurate forecasts.

We then check the significance of the differences on the medians
via the
Bayesian signed-rank test \cite{benavoli2014bayesian}, which
is  a Bayesian  counterpart of the Wilcoxon signed-rank test. It returns posterior probabilities instead of the p-value.
An advantage of this test over the frequentist one is that we can set
a region of practical equivalence (rope) between the two algorithms being compared.
When comparing algorithms A and B, the test
returns three posterior probabilities:
the probability of the two algorithms being practically equivalent,
i.e, the probability of the median difference belonging to the rope;
the probability of A being significantly better than B,
and vice versa.
As already pointed out, better means lower MAE, lower CRPS, higher LL.
We considered a rope of $\pm$0.01 on each indicator, similarly to \cite{benavoli2017time}.
We consider as significant the differences in which the probability of an algorithm
being better than another is at least 95\%.
The improvement yielded by the GP over the competitors are
significant in most cases; see the starred entries in Tab.\ref{tab:results-m}.
When the median of some competitor was better than that of the GP, the difference was not statistically significant.

\begin{figure}[!htp]
	\centering
	\includegraphics[width=0.49\textwidth]{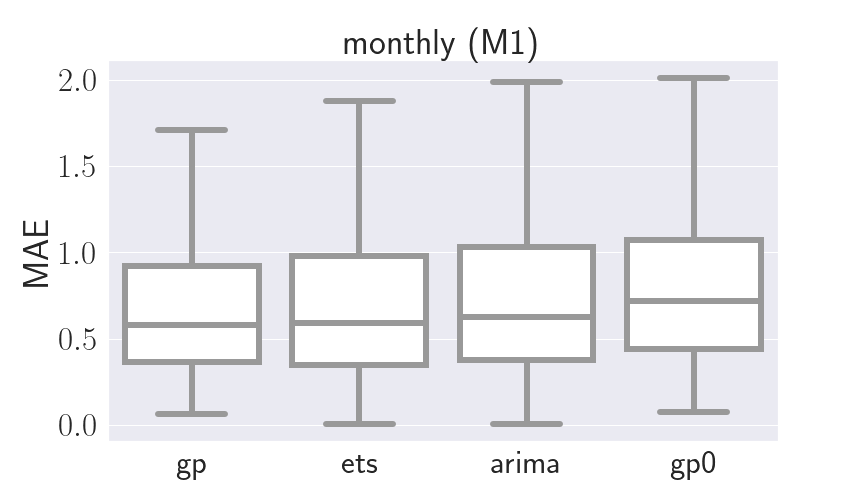}
	\includegraphics[width=0.49\textwidth]{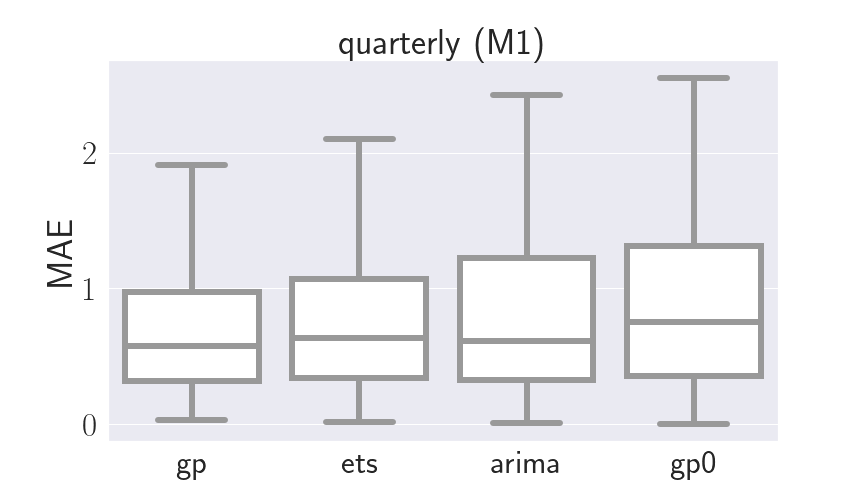}\\
	\bigskip
	\includegraphics[width=0.49\textwidth]{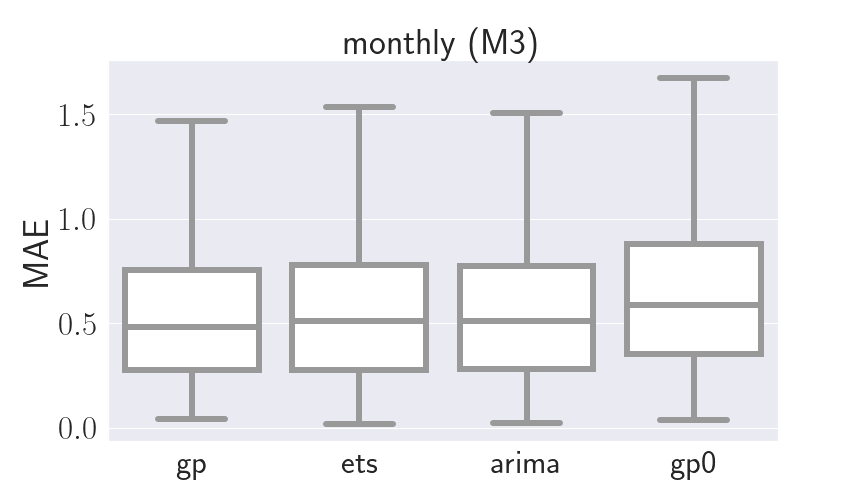}
	\includegraphics[width=0.49\textwidth]{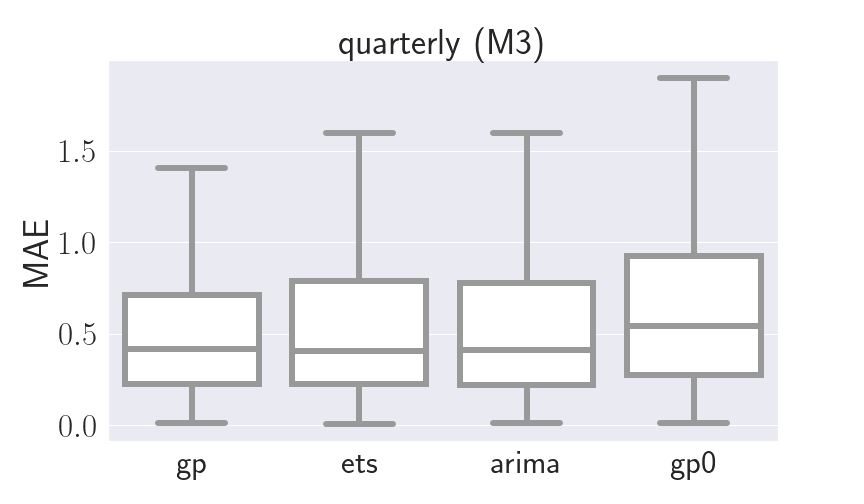}\\
	\bigskip
	\caption{Distribution of MAE on the monthly and quarterly time series of the M1 and M3 competition.}
	\label{fig:bplot-m}
\end{figure}

The improvement is not only on the medians, but it also involve the distribution across time series, as shown by the boxplots of MAE
(Fig \ref{fig:bplot-m}).
Similar results hold also for the distribution of the other indicators, which we do not show for reasons of space.

\section{Dealing with multiple seasonalities}\label{sec:energy}
We then test the versatility of our GP model, by considering
time series with multiple seasonalities.
We consider
the electricity data set\footnote{\url{https://archive.ics.uci.edu/ml/datasets/ElectricityLoadDiagrams20112014}}, which contains
370 time series regarding electricity demand  for different Portoguese households.

Each time series covers the period January 2011 - September 2015 with a sampling frequency of 15 mins,  totaling 140k points.
To have a more manageable data set we aggregate the data to
6-hours steps.
We consider a training set containing of 250 days
(1000 points) and
a test set of 10.5 days (42 steps).

Such time series have a daily and a weekly seasonal pattern.
This can be addressed by approaches
which model seasonality
using Fourier terms.
For instance TBATS \cite{de2011forecasting}
introduces Fourier terms within an exponential smoothing state space model.
Prophet \cite{taylor2018forecasting} is a decomposable Bayesian time series model, whose final forecast is the sum of different functions, which account for different effects. The seasonality function is is modeled by Fourier terms. Prophet however is not very effective on time series with simpler seasonality, such as monthly and quarterly time series, as we have already seen.

We adapt the kernel composition by adding a second periodic kernel:
\begin{center}
	\centering
	K = PER$_w$  + PER$_d$ +  LIN + RBF + SM$_1$ + SM$_2$,
\end{center}
where PER$_w$ and
PER$_d$  represents respectively  the weekly and the daily pattern.
We thus set the  period of PER$_w$ to $\frac{1}{52.18}$
and the period of PER$_d$
to $\frac{1}{(365.25)}$
As in previous experiments, we standardize time series
and one year corresponds to time
increasing of one unit. We can keep unchanged the priors.

\setlength{\tabcolsep}{8pt}
\renewcommand\arraystretch{1.5}
\begin{table}[!htp]
	\centering
	\begin{tabular}{@{}lrrr@{}}
		\toprule
		& \multicolumn{1}{c}{GP$\,\,\,\,\,\,\,$} & \multicolumn{1}{c}{Tbats$\,\,\,\,\,$} & \multicolumn{1}{c}{\small{Prophet}} \\ \midrule
		\rowcolor{Gray}
		MAE  & \textbf{0.26$\,\,$}          & 0.30$^*$                      & 0.29$^*$                        \\
		CRPS & \textbf{0.19$\,\,$}          & 0.23$^*$                      & 0.21$^*$                       \\
		\rowcolor{Gray}
		LL   & \textbf{-0.37$\,\,$}         & -0.60$^*$                     & -0.49$^*$                       \\ \bottomrule
	\end{tabular}
	\caption{Median results on the electricy data sets (370 time series). Starred results imply statistical significance (Bayesian signed rank test).}
	\label{tab:elec-medians}
\end{table}

\begin{figure}[!htp]
	\centering
	\includegraphics[width=0.5\textwidth]{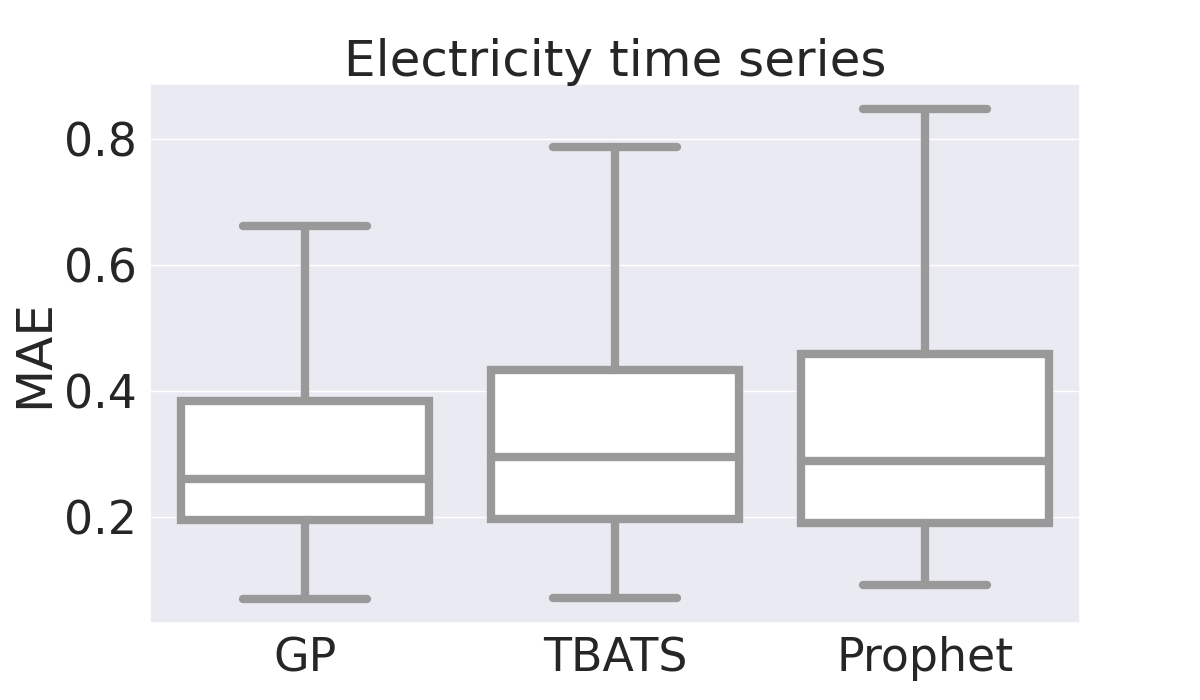}
	\caption{MAE Boxplot on the 370 electricity time series.}
	\label{fig:bplot-electricity}
\end{figure}

In table \ref{tab:elec-medians} we report the median results across the 370 time series;
the GP delivers the best  performance on all indicators.
The GP model compares favorably to the competitors also as for
the distribution of the MAE (Fig. \ref{fig:bplot-electricity})
across time series.

\section{Code and replicability}
We submit
as supplementary material  our code and the data which allow
replicating the experiments on M1 and M3 time series.
Our implementation is based on the GPy library
\cite{gpy2014}.

\section{Conclusions}
As far as we know, these are the best results obtained so far in automatic forecasting with Gaussian processes.
Our model is competitive  with the best time series models on different types of time series: monthly, quarterly and time series with multiple seasonalities.

The model is fast to train, at least on time series containing less than 500 data points. Recent computational advances with GPs in time series \cite{solin2014explicit,ambikasaran2015fast} could allow the application of
our methodology also to  time series thousands of observations.

Our GP model yields both good point forecast and a reliable quantification of the uncertainty, as shown by the CRPS and LL indicators.
It is thus an interesting candidate for problems of hierarchical forecasting \cite{wickramasuriya2019optimal},
which require forecasts with a sound quantification of the uncertainty.

Due to the general properties of the GP, the model can be learned also from irregularly sampled or incomplete time series.

\end{document}